\documentclass[a4paper]{article}
\usepackage{latexsym}
\usepackage{colacl}
\usepackage{graphics}
\usepackage{epic}
\usepackage{eepic}
\usepackage{amsmath}
\usepackage{amssymb}
\usepackage{pstricks}
\usepackage{pst-node}

\newenvironment{definition}[1][Definition]{\begin{trivlist}
\item[\hskip \labelsep {\bfseries #1}]}{\end{trivlist}}

\newcommand{\qed}{\nobreak \ifvmode \relax \else
      \ifdim\lastskip<1.5em \hskip-\lastskip
      \hskip1.5em plus0em minus0.5em \fi \nobreak
      \vrule height0.75em width0.5em depth0.25em\fi}

\def\Pr{\mbox{\bf \sf P}}
\newcommand{\bigjoin}{\overset{\ }{\raisebox{-1mm}{\mbox{\huge $\Join$}}}}  

\newcommand{\order}[1]{${\mathcal{O}}(#1)$}
\newcommand{\sent}[1]{{`#1'}}

\newcommand{\tabbox}[2]
 { \rnode{#1}{\psframebox{\begin{tabular}{@{}c@{}} #2 \end{tabular}}} }

\newcommand{\tuple}[1]{ \langle #1 \rangle }

\newcommand{\vareps}{ \varepsilon }
\newcommand{\wild}{\_\,}

\title { Interleaved semantic interpretation in environment-based parsing%
\footnotemark[1] }

\author
 {
   William Schuler \\
   Computer and Information Science Dept. \\
   University of Pennsylvania \\
   Philadelphia, PA 19103 \\
   {\tt schuler@linc.cis.upenn.edu}
 }

\begin{document}

\maketitle

\def\thefootnote{\fnsymbol{footnote}}
\footnotetext[1]
 {
 The author would like to thank David Chiang, Karin Kipper, and Alexander Koller, 
 as well as the anonymous reviewers for comments on this material.
 This work was partially supported by NSF IIS-9900297 and DARPA N66001-00-1-8915.
 }

\def\thefootnote{\arabic{footnote}}

\begin{abstract}
This paper extends a polynomial-time parsing algorithm that 
resolves structural ambiguity in input sentences by calculating and comparing 
the denotations of rival constituents, given some model of the application 
environment \cite{schuler01}. 
The algorithm is extended to incorporate a full set of logical operators,
including quantifiers and conjunctions, 
into this calculation without increasing the complexity of the overall 
algorithm beyond polynomial time, both in terms of the length of the input and 
the number of entities in the environment model.
\end{abstract}

\section{Introduction}


The development of speaker-independent mixed-initiative speech 
interfaces, in which users not only answer questions but also ask
questions and give instructions, is currently limited by the
inadequacy of existing corpus-based disambiguation techniques.
This paper explores the use of semantic and pragmatic information, in
the form of the entities and relations in the interfaced application's
run-time environment, as an additional source of information to guide
disambiguation.

In particular, this paper extends an existing parsing algorithm that
calculates and compares the denotations of rival 
parse tree constituents in order to resolve structural ambiguity in 
input sentences \cite{schuler01}.
The algorithm is extended to incorporate a full set of logical operators 
into this calculation 
so as to 
improve the accuracy of the resulting denotations -- and thereby improve 
the accuracy of parsing -- 
without increasing the complexity of the overall algorithm beyond 
polynomial time (both in terms of the length of the input 
and the number of entities in the environment model).
This parsimony is achieved by localizing certain kinds of semantic relations during 
parsing, particularly those between quantifiers and their restrictor and 
body arguments
(similar to the way dependencies between predicate and argument head words 
are localized in lexicalized formalisms such as tree adjoining grammars),
in order to avoid calculating exponential higher-order denotations for 
expressions like generalized quantifiers. 

\section{Basic algorithm}
\label{sect:basic}

This section describes the basic environment-based parser \cite{schuler01}
which will be extended in Section~\ref{sect:extend}.
Because it will crucially rely on the denotations (or interpretations) 
of proposed constituents 
in order to guide disambiguation, the parser
will be defined on categorial grammars \cite{ajdukiewics35,barhillel53}, 
whose categories 
all have well defined types and worst-case denotations.
These categories are drawn from a minimal set of symbols ${\mathcal C}$ such that:
\[
\begin{array}{l}
\mathrm{NP} \in {\mathcal C} \text{ and } \mathrm{S} \in {\mathcal C}, \\
\text{if } \gamma, \delta \in {\mathcal C} 
  \text{ then } \gamma/\delta \in {\mathcal C}
  \text{ and } \gamma\backslash\delta \in {\mathcal C}.
\end{array}
\]
Intuitively, the category $\mathrm{NP}$ describes a noun phrase and the category
$\mathrm{S}$ describes a sentence,
and the complex categories $\gamma/\delta$ and $\gamma\backslash\delta$ describe 
`a~$\gamma$ lacking a~$\delta$ to the right' and 
`a~$\gamma$ lacking a~$\delta$ to the left' respectively; 
so for example $\mathrm{S}\backslash\mathrm{NP}$ 
would describe a declarative verb phrase lacking an $\mathrm{NP}$ 
subject to its left in the input.

The type $T$ and worst-case (most general) denotation $W$ of each possible category are 
defined below, given a set of entities~${\mathcal E}$ as an environment:
\[
\begin{array}{@{}l@{\ \ \ \ \ }l@{}}
T(\mathrm{S})                       = t : \text{truth value}
 & W(\mathrm{S})                      = \{\mathrm{TRUE},\mathrm{FALSE}\} \\
T(\mathrm{NP})                      = e : \text{entity}
 & W(\mathrm{NP})                     = {\mathcal E} \\
T(\mathrm{\gamma/\delta})           = \langle T(\delta), T(\gamma) \rangle
 & W(\mathrm{\gamma/\delta})          = W(\delta) \times W(\gamma) \\
T(\mathrm{\gamma\backslash\delta})  = \langle T(\delta), T(\gamma) \rangle
 & W(\mathrm{\gamma\backslash\delta}) = W(\delta) \times W(\gamma) \\
\end{array}
\]
The denotation $D$ of any proposed constituent is constrained to be 
a subset of the worst-case denotation $W$ of the constituent's category; 
so a constituent of category $\mathrm{NP}$ would denote a set of entities, 
$\{ e_1, e_2, \dots \}$, and a constituent of category 
$\mathrm{S}\backslash\mathrm{NP}$ would denote a set of entity $\!\times\!$ 
truth value pairs, $\{ \langle e_1,\mathrm{TRUE} \rangle, 
\langle e_2,\mathrm{FALSE} \rangle, \dots \}$.
Note that no denotation of a constituent can contain more than 
\order{|{\mathcal E}|^v} different elements, where $v$ is a valency measure of 
the number of $\mathrm{NP}$ symbols occurring within the constituent's category.

This paper will use the following definition of a categorial grammar (CG):
\begin{definition}
A {\em categorial grammar} $G$ is a formal grammar $(N,\Sigma,P)$ such that:

\begin{itemize}

  \item $\Sigma$ is a finite set of words $w$;

  \item $P$ is a finite set of productions containing: \\
        \begin{tabular}{@{}ll}
        $\gamma \rightarrow w$
             & for all $w \!\in\! \Sigma$, with $\gamma \in {\mathcal C}$, \\
        $\gamma \rightarrow \gamma/\delta \ \ \delta$ 
             & for every rule $\gamma/\delta \rightarrow \dots$ in $P$, \\
        $\gamma \rightarrow \delta \ \ \gamma\backslash\delta$
             & for every rule $\gamma\backslash\delta \rightarrow \dots$ in $P$, \\
        \multicolumn{2}{@{}l}{ and nothing else; } \\
        \end{tabular}

  \item $N$ is the nonterminal set
        $\{ \gamma \mid \gamma \rightarrow \dots \ \in P \}$.

\end{itemize}
\end{definition}
and the following deductive parser,%
\footnote{Following Shieber et al.~\shortcite{shieberetal95}.}
which will be extended later to handle 
a richer set of semantic operations.
The parser is defined with:
%
\begin{itemize}

  \item constituent chart items $[i,j,\gamma]$ 
        drawn from ${\mathbb I}_0^n \times {\mathbb I}_0^n \times N$, 
        indicating that positions $i$ through $j$ in the input can 
        be characterized by category $\gamma$;

  \item a lexical item $[i,j,\gamma]$
        for every rule $\gamma \rightarrow w \in P$ 
        if $w$ occurs between positions $i$ and $j$ in the input; 

  \item a set of rules of the form:

        $\frac{[i,k,\gamma/\delta] \ [k,j,\delta]}{[i,j,\gamma]}$
        for all $\gamma \rightarrow \gamma/\delta \ \ \delta \in P, 
                 \ i,j,k \in {\mathbb I}_0^n$,

        $\frac{[k,j,\gamma\backslash\delta] \ [i,k,\delta]}{[i,j,\gamma]}$
        for all $\gamma \rightarrow \delta \ \ \gamma\backslash\delta \in P, 
                 \ i,j,k \in {\mathbb I}_0^n$.
\end{itemize}
%
and can recognize an $n$-length input as a constituent 
of category~$\gamma$ (for example, as an~$\mathrm{S}$) if it can deduce
the chart item $[0,n,\gamma]$.

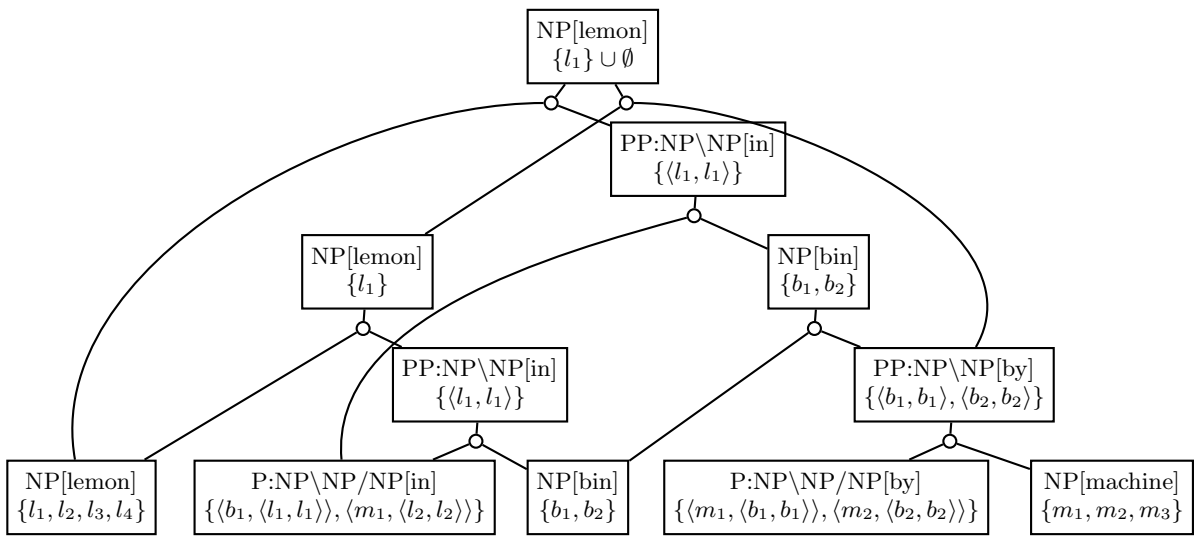
\begin{figure*}[htbp]
\begin{center}
\small
\begin{pspicture}(16,7)
\rput(1.2,0.5) { \tabbox{p1}  { NP[lemon] \\ $\{l_1,l_2,l_3,l_4\}$}}
\rput(4.7,0.5) { \tabbox{p2}  { P:NP$\backslash$NP/NP[in] \\ 
   $\{\langle b_1,\langle l_1,l_1 \rangle \rangle,
      \langle m_1,\langle l_2,l_2 \rangle \rangle\}$}}
\rput(7.8,0.5) { \tabbox{p3}  { NP[bin] \\ $\{b_1,b_2\}$}}
\rput(11.1,0.5) { \tabbox{p4} { P:NP$\backslash$NP/NP[by] \\ 
   $\{\langle m_1,\langle b_1,b_1 \rangle \rangle,
      \langle m_2,\langle b_2,b_2 \rangle \rangle\}$}}
\rput(14.9,0.5) { \tabbox{p5} { NP[machine] \\ $\{m_1,m_2,m_3\}$}}
\rput(6.5,2) { \tabbox{p23}   { PP:NP$\backslash$NP[in] \\ $\{\langle l_1,l_1\rangle\}$}}
\rput(12.8,2) { \tabbox{p45}  { PP:NP$\backslash$NP[by] \\
   $\{\langle b_1,b_1 \rangle, \langle b_2,b_2 \rangle\}$}}
\rput(5,3.5) { \tabbox{p13}   { NP[lemon] \\ $\{l_1\}$}}
\rput(11,3.5) { \tabbox{p35}  { NP[bin] \\ $\{b_1,b_2\}$}}
\rput(9.4,5) { \tabbox{p25}   { PP:NP$\backslash$NP[in] \\ $\{\langle l_1,l_1\rangle\}$}}
\rput(8,6.5) { \tabbox{p15}   { NP[lemon] \\ $\{l_1\} \cup \emptyset$}}
\cnode(6.5,1.25){1mm}{p23a}
  \ncline{p2}{p23a} \ncline{p3}{p23a}
  \ncline{p23a}{p23}
\cnode(12.8,1.25){1mm}{p45a}
  \ncline{p4}{p45a} \ncline{p5}{p45a}
  \ncline{p45a}{p45}
\cnode(5,2.75){1mm}{p13a}
  \ncline{p1}{p13a} \ncline{p23}{p13a}
\cnode(11,2.75){1mm}{p35a}
  \ncline{p3}{p35a} \ncline{p45}{p35a}
  \ncline{p13a}{p13}
  \ncline{p35a}{p35}
\cnode(9.4,4.25){1mm}{p25a}
  \nccurve[angleA=95,angleB=195]{p2}{p25a} \ncline{p35}{p25a}
  \ncline{p25a}{p25}
\cnode(8.5,5.75){1mm}{p15a}
  \ncline{p13}{p15a} \nccurve[angleA=60,angleB=0]{p45}{p15a}
\cnode(7.5,5.75){1mm}{p15b}
  \nccurve[angleA=100,angleB=180]{p1}{p15b} \ncline{p25}{p15b}
  \ncline{p15a}{p15} \ncline{p15b}{p15}
\end{pspicture}
\vspace{-1mm}
\end{center}
\caption{Denotation-annotated forest for \sent{lemon in bin by machine.}}
\label{fig:forest}
\end{figure*}

This parser can be implemented in a dynamic programming algorithm,
using the recursive function:
\[
F(x) = \!\!\!\!\!
  \underset{a_1 ...\, a_k\, s.t.\, \frac{a_1 ...\, a_k}{x}}{\bigvee} \!\!
    \ \ \overset{k}{\underset{i=1}{\bigwedge}}
      \ \ F(a_i)
\]
(where $x, a_1 \dots\, a_k$ are proposed constituents drawn from 
${\mathbb I}_0^n \!\times\! {\mathbb I}_0^n \!\times\! N$,
$\bigvee_{\emptyset}=\mathrm{FALSE}$, and $\bigwedge_{\emptyset}=\mathrm{TRUE}$),
by recording the result of every
recursive sub-call to $F(x)$ in a chart, then consulting this chart on
subsequent calls to $F(x)$ for the same $x$ constituent.
\footnote{Following Goodman \shortcite{goodman99}.} 
Since the indices in every rule's antecedent constituents~$a_1 \dots\, a_k$ 
each cover smaller spans than those in the consequent~$x$, the algorithm will 
not enter into an infinite recursion; 
and since there are only $n^2|N|$ different values of~$x$, and only
$2n$ different rules that could prove any consequent~$x$
(two rule forms for $/$ and $\backslash$, each with $n$ different values of~$k$), 
the algorithm runs in polynomial time: \order{n^3|N|}.
The resulting chart can then be annotated with back pointers 
to produce a polynomial-sized shared forest representation of all possible 
grammatical trees \cite{billotlang89}.

Traditional corpus-based parsers select preferred trees from such forests 
by calculating Viterbi scores for each proposed constituent, 
according to the recursive function:
\[
S_V(x) = \!\!\!\!\!\!\!\!
  \underset{a_1 ...\, a_k\, s.t.\, \frac{a_1 ...\, a_k}{x}}{\mbox{max}} \!\!
     \left( \overset{k}{\underset{i=1}{\prod}}
              S_V(a_i) \right)
     \cdot \Pr ( a_1 ...\, a_k \mid x ) 
\]
These scores can be calculated in polynomial time, 
using the same dynamic programming algorithm as that described for parsing.
A tree can then be selected, from the top down, by expanding the highest-scoring 
rule application for each constituent.

The environment-based parser described here
uses a similar mechanism to select preferred trees, 
but the scores are based on the presence or absence of entities in the 
denotation (interpretation) of each proposed constituent:%
\footnote{Here, the score is simply equal to the number of non-empty constituents 
in an analysis, but other metrics are possible.}
\[
S_D(x) = \!\!\!\!\!\!\!\!\!\!
  \underset{a_1\dots\, a_k\, s.t.\, \frac{a_1\dots\, a_k}{x}}{\mbox{max}} \!\!
    \left( \overset{k}{\underset{i=1}{\sum}}
           S_D(a_i) \right)
    + \begin{cases} 1 & \!\!\text{if } D(x)\!\neq\!\emptyset \\
                    0 & \!\!\text{otherwise}  \end{cases}
\]
where the denotation $D(x)$ of a proposed constituent $x$ is 
calculated using another recursive function:
\[
D(x) = \!\!\!\!\!\!\!\!\!\!\!
  \underset{a_1\dots\, a_k\, s.t.\, \frac{a_1\dots\, a_k}{x}}{\bigcup} \!\!\!
    \left( \!\pi\! \operatornamewithlimits{\bigjoin}\limits_{i=1}^{k} \!
           D(a_i) \right)
    \!\Join\! \begin{cases} R(x)               & \!\!\!\text{if } k=0 \\
                            \{\langle\rangle\} & \!\!\!\text{otherwise}  \end{cases}
\]
in which $R(x)$ is a lexical relation defined for each axiom $x$ of category $\gamma$ 
equal to some subset of $\gamma$'s worst-case denotation $W(\gamma)$, as defined above.%
\footnote{So a lexical relation for the constituent \sent{lemon} 
of category $\mathrm{NP}$ would contain all and only the lemons in the
environment, and a lexical relation for the constituent \sent{falling} 
of category $\mathrm{S}\backslash\mathrm{NP}$ would contain a mapping 
from every entity in the environment to some truth value (TRUE if that 
entity is falling, FALSE otherwise): 
e.g.~$\{\tuple{lemon_1,TRUE},\tuple{lemon_2,FALSE},\dots \}$.}
The operator $\Join$ is natural (relational) join on the fields of its operands:
\[
A \!\Join\! B = \{ \langle e_1 \!... e_{max(a,b)} \rangle \mid
                   \langle e_1 \!... e_a \rangle \!\in\! A, 
                   \langle e_1 \!... e_b \rangle \!\in\! B \}
\]
where $a,b\geq0$; and $\pi$ is a projection that removes the first element 
of the result (corresponding the most recently discharged argument of the 
head or functor category):
\[
\pi A = \{ \langle e_2   ... e_a \rangle \mid
           \langle e_1 \!... e_a \rangle \!\in\! A \}
\]
This interleaving of semantic evaluation and parsing for the purpose of 
disambiguation has much in common with that of Dowding et al.~\shortcite{dowdingetal94}, 
except that in this case, constituents are not only semantically type-checked, but 
are also fully interpreted each time they are proposed.

Figure~\ref{fig:forest} shows a sample denotation-annotated forest for the phrase
\sent{the lemon in the bin by the machine}, using the lexicalized grammar:
\[
\begin{array}{r@{\ }c@{\ }l}
\text{lemon, bin, machine} &:& \mathrm{NP} \\
\text{the}                 &:& \mathrm{NP}/\mathrm{NP} \\
\text{in, by}              &:& \mathrm{NP}\backslash\mathrm{NP}/\mathrm{NP} \\
\end{array}
\]
in which the denotation of each constituent (the set in each
rectangle) is calculated using a join on the denotations of each pair
of constituents that combine to produce it.
In this example, the right-branching tree would be preferred 
because the denotation resulting from the composition at the root 
of the other tree would be empty.

Since this use of the join operation is linear on the sum of the cardinalities 
of its operands, and since the denotations of the categories in a 
grammar~$G$ are bounded in cardinality by \order{|{\mathcal E}|^v} 
where $v$ is the maximum valency of the categories in~$G$, 
the total complexity of the above algorithm can be shown to be 
\order{n^3|{\mathcal E}|^v}: polynomial not only on the length of the input $n$, 
but also on the size of the environment~${\mathcal E}$ \cite{schuler01}.



\section{Extended algorithm}
\label{sect:extend}

The above algorithm works well for attaching ordinary complements and modifiers,
but as a semantic theory it is not sufficiently expressive to produce correct 
denotations in all cases.
For example, 
the lexical relations defined above 
are insufficient to represent quantifiers like \sent{no} 
(using category $\mathrm{NP}/\mathrm{NP}$) in the phrase 
\sent{the boy with no backpack.}%
\footnote{Assigning the identity relation 
$\{ \langle e_1,e_1 \rangle, \langle e_2,e_2 \rangle, \dots \}$ 
to the quantifier would incorrectly yield the set of boys {\em with} a backpack 
as a denotation for the full noun phrase;
and assigning the converse relation (from each entity in the environment 
to every other entity $\{ \langle e_1,e_2 \rangle, \langle e_1,e_3 \rangle, \dots \}$)
would incorrectly yield the set of boys with anything that is not a backpack.}
A similar problem occurs with conjunctions;
for example, the word \sent{and}
(using category $\mathrm{NP}\backslash\mathrm{NP}/\mathrm{NP}$) in the phrase 
\sent{the child wearing glasses and blue pants}, also cannot be properly 
represented as a lexical relation.%
\footnote{The identity relation 
$\{ \langle e_1,e_1,e_1 \rangle, \langle e_2,e_2,e_2 \rangle, \dots \}$, 
which yields a correct interpretation in verb phrase conjunction,
would yield an incorrect denotation for the noun phrase \sent{glasses and blue pants,}
containing only entities which are at once both glasses and pants.}
This raises the question: how much expressivity can be allowed in a shared semantic 
interpretation without exceeding the tractable parsing complexity 
necessary for practical environment-based parsing?

In traditional categorial semantics \cite{montague73,barwisecooper81,keenanstavi86}
quantifiers and noun phrase conjunctions denote higher-order relations: 
that is, relations between whole sets of entities instead of just between individuals.
Under this interpretation, a quantifier like \sent{no} would denote a set of 
pairs $\{ \langle A_1,B_1 \rangle, \langle A_2,B_2 \rangle, \dots \}$ where 
each $A_i$ and $B_i$ are disjoint subsets of ${\mathcal E}$, corresponding to 
an acceptable pair of restrictor and body sets satisfying the 
quantifier \sent{no}.
Unfortunately, since the cardinalities of these higher-order denotations can 
be exponential on the size of the environment ${\mathcal E}$ 
(there are $2^{|{\mathcal E}|}$ possible subsets of ${\mathcal E}$ 
and $2^{2|{\mathcal E}|}$ possible combinations of two such subsets), 
such an approach would destroy the polynomial complexity of the environment-based 
parsing algorithm. 


However, if the number of possible higher-order functions is restricted to 
a finite set (say, to some subset of words in a lexicon), it becomes tractable 
to store them by name rather than by denotation (i.e.~as sets).
Such function can then discharge 
all their first-order arguments in a single derivational step to produce a first-order 
result, in order to avoid generating or evaluating any higher-order partial results.
Syntactically, this would be analogous to composing a quantifier with both a noun 
phrase restrictor and a body predicate (e.g.~a verb or verb phrase) at the same time, 
to produce another first-order predicate (e.g.~a verb phrase or sentence).
Since a generalized quantifier function merely counts and compares the cardinalities 
of its arguments in a linear time operation, this analysis provides a tractable 
shortcut to the exponential calculations required in the conventional analysis.

Note that this analysis by itself does not admit productive modification of 
quantifiers (because their functions are drawn from some finite set) or of quantified 
noun phrases (because they are no longer derived as a partial result).
This causes no disruption to the attachment of non-conjunctive modifiers, because 
ordinary syntactic modifiers of quantifier constituents are seldom productive 
(in the sense that their composition does not yield functions outside some finite set), 
and syntactic modifiers of NP constituents usually only modify the restrictor set of 
the quantifier rather than the entire quantified function, and can therefore safely 
be taken to attach below the quantifier, to the unquantified NP.

But this is not true in cases involving conjunction.
Conjoined quantifiers, like \sent{some but not all,} cannot always be defined 
using a single standard lexical function; and conjunctions of quantified noun phrases, 
like \sent{one orange and one lemon}, cannot be applied to unquantified subconstituents 
(syntactically, because this would fail to subsume the second quantifier, and 
semantically, because it is not the restrictor sets which are conjoined).
Keenan and Stavi \shortcite{keenanstavi86} model conjunctions of quantifiers 
and quantified noun phrases using lattice operations on higher-order sets, 
but as previously stated, these higher-order sets preclude tractable 
interleaving of semantic interpretation with parsing.

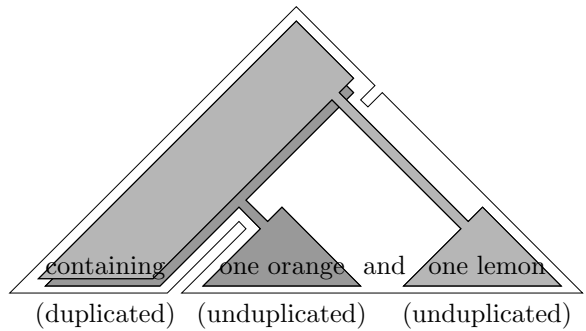
\begin{figure}
\begin{center}
\setlength{\unitlength}{.0125in}
\begin{picture}(241,122)(0,0)
\path(120,120)(0,0)(66,0)
        (96,30)(99,27)(72,0)
        (240,0)(156,84)(150,78)
        (147,81)(153,87)(120,120)
\texture{55888888 88555555 5522a222 a2555555 55888888 88555555 552a2a2a 2a555555 
        55888888 88555555 55a222a2 22555555 55888888 88555555 552a2a2a 2a555555 
        55888888 88555555 5522a222 a2555555 55888888 88555555 552a2a2a 2a555555 
        55888888 88555555 55a222a2 22555555 55888888 88555555 552a2a2a 2a555555 }
\shade\path(120,108)(15,3)(63,3)
        (96,36)(105,27)(81,3)
        (147,3)(114,36)(108,30)
        (99,39)(144,84)(120,108)
\path(120,108)(15,3)(63,3)
        (96,36)(105,27)(81,3)
        (147,3)(114,36)(108,30)
        (99,39)(144,84)(120,108)
\texture{88555555 55000000 555555 55000000 555555 55000000 555555 55000000 
        555555 55000000 555555 55000000 555555 55000000 555555 55000000 
        555555 55000000 555555 55000000 555555 55000000 555555 55000000 
        555555 55000000 555555 55000000 555555 55000000 555555 55000000 }
\shade\path(120,114)(12,6)(60,6)
        (135,81)(189,27)(165,3)
        (231,3)(198,36)(192,30)
        (138,84)(144,90)(120,114)
\path(120,114)(12,6)(60,6)
        (135,81)(189,27)(165,3)
        (231,3)(198,36)(192,30)
        (138,84)(144,90)(120,114)
\put(40,8){\makebox(0,0){\smash{containing}}}
\put(40,-12){\makebox(0,0){\smash{(duplicated)}}}
\put(114,8){\makebox(0,0){\smash{one orange}}}
\put(114,-12){\makebox(0,0){\smash{(unduplicated)}}}
\put(157,8){\makebox(0,0){\smash{and}}}
\put(200,8){\makebox(0,0){\smash{one lemon}}}
\put(200,-12){\makebox(0,0){\smash{(unduplicated)}}}
\end{picture}
\end{center}

\vspace{2mm}

\caption{Duplicated verb in NP conjunction.}
\label{fig:qpconj}
\end{figure}

The solution proposed here is to treat each quantifier or quantified noun phrase 
conjunction as an elliptical conjunction of two complete first-order 
predicates (e.g.~verb phrases or sentences), each subsuming a different quantifier 
and noun phrase restrictor (in the case of NP conjunction), but sharing or 
{\em duplicating} a common body predicate.
This analysis requires multiple components to keep track of the duplicated material 
above the conjunction, but as long as the number of components is bounded, the 
polynomial complexity of the parsing algorithm is retained.%
\footnote{Dahl and McCord \shortcite{dahlmccord83} propose a similar duplication 
mechanism to produce appropriate semantic representations for NP and other 
conjunctions, but for different reasons.} 

Figure~\ref{fig:qpconj} shows a duplicated verb predicate in the
derivation of an NP conjunction.
The conjoined constituents (the shaded regions in the figure) are each 
composed of two components:
one for the NP itself, containing the quantifier and the restrictor predicate, 
and one for the verb which supplies the body predicate of the quantifier.
Since the conjoined constituents both correspond to complete quantifier 
expressions with no unsatisfied first-order arguments, their categories are 
that of simple first-order predicates (they are each complete verb phrases 
in essence: \sent{containing one orange} and \sent{containing one lemon}).
The conjunction then forms a larger constituent of the same form 
(the unshaded outline in the figure), with a lower 
component containing the conjoined constituents' NP components 
concatenated in the usual way, and an upper component in which the 
conjoined constituents' non-NP components are identified or overlapped. 
If the duplicated components do not cover the same string yield, 
the conjunction does not apply.

Note that, since they are only applied to ordinary first-order predicates 
(e.g.~sentences or verb phrases) in this analysis, 
conjunctions can now safely be assigned the familiar truth-functional denotations 
in every case.%
\footnote{e.g.~for the word \sent{and}:
$\{\tuple{...\mathrm{TRUE},...\mathrm{TRUE},...\mathrm{TRUE}},$
$\tuple{..\mathrm{TRUE},..\mathrm{FALSE},..\mathrm{FALSE}},$
$\tuple{..\mathrm{FALSE},..\mathrm{TRUE},..\mathrm{FALSE}},$
$\tuple{..\mathrm{FALSE},..\mathrm{FALSE},..\mathrm{FALSE}}\}$
}
Also, since the resulting constituent has the same number of components 
as the conjoined constituents, there is nothing to prevent its 
use as an argument in subsequent conjunction operations.

A sample multi-component analysis for quantifiers is shown below, 
allowing material to be duplicated both to the left and to the right 
of a conjoined NP:
\[
\begin{array}{r@{\ }c@{\ }l}
\text{some,all,no,etc.} &:& X\backslash\mathrm{NP}_q \cdot
                              \mathrm{NP}_q\backslash\mathrm{NP}_q \cdot
                                \mathrm{NP}_q/\mathrm{NP}_\epsilon  \\
                        & & X/\mathrm{NP}_q \cdot
                              \mathrm{NP}_q\backslash\mathrm{NP}_q \cdot
                                \mathrm{NP}_q/\mathrm{NP}_\epsilon  \\
\end{array}
\]
The lexical entry for a quantifier can be split in this way into a number of 
components, the last (or lowest) of which is not duplicated in conjunction while 
others may or may not be.
These include a component for the quantifier $\mathrm{NP}_q/\mathrm{NP}_\epsilon$ 
(which will ultimately also contain a noun phrase restrictor of category 
$\mathrm{NP}_\epsilon)$, 
a component for restrictor PPs and relative clauses of category 
$\mathrm{NP}_q\backslash\mathrm{NP}_q$ that are attached above the quantifier 
and duplicated in the conjunction, 
and a component for the body (a verb or verb phrase or other predicate) 
of category $X\backslash\mathrm{NP}_q$ or $X/\mathrm{NP}_q$.
The subscript $q$ specifies one of a finite set of quantifiers, and the subscript 
$\epsilon$ indicates an unquantified NP.

The deductive parser presented in Section~\ref{sect:basic} can now be 
extended by incorporating sequences of recognized and 
unrecognized components into the constituent chart items.
As constituents are composed, components are shifted from the 
unrecognized sequence $\gamma_1\cdots\gamma_c$ to the recognized sequence 
$\tuple{i_1,j_1,\gamma_1}\cdots\tuple{i_c,j_c,\gamma_c}$, 
until the unrecognized sequence is empty.

The extended parser is defined with:
\begin{itemize}

  \item chart items of the form
        $[i,j,\Delta,\Sigma]$, where
        $\Delta$ is a sequence of unrecognized components~$\gamma$, 
        $\Sigma$ is a sequence of recognized components~$\tuple{a,b,\gamma}$, 
        and $i,j,k,a,b,c$ are indices in the input.
        Each item $[i,j,\Delta\cdot\gamma,
        \tuple{i_1,j_1,\gamma_1}\cdots\tuple{i_c,j_c,\gamma_c}]$
        indicates that the span from $i$ to $j$ in the input can be characterized 
        by the categories $\gamma_1$ through $\gamma_c$ at positions $i_1$ to $j_1$ 
        through $i_c$ to $j_c$ respectively, so that if these spans are concatenated 
        in whatever order they occur in the input string, they form a grammatical 
        constituent of category~$\gamma$ with unrecognized components $\Delta$.

  \item a lexical item $[i,j,\gamma,\tuple{i,j,\gamma}]$
        for every rule $\gamma \rightarrow w \in P$ 
        if $w$ occurs between positions $i$ and $j$ in the input; 

  \item a set of rules for all $i,j,k,a,b,c \in {\mathbb I}_0^n$ as below.

   Two rules to invoke left and right function application
   to an existing component:
   \begin{center}
   $\frac{[i,k,\gamma/\delta,\tuple{i,k,\gamma/\delta}] 
          \ [k,j,\Delta\cdot\delta,\tuple{k,b,\delta/\vareps}\cdot\Sigma]}
         {[i,j,\Delta\cdot\gamma,\tuple{i,b,\gamma/\vareps}\cdot\Sigma]}
      {\scriptstyle \gamma \rightarrow \gamma/\delta \ \delta \in P}$,

   $\frac{[k,j,\gamma\backslash\delta,\tuple{k,j,\gamma\backslash\delta}] 
          \ [i,k,\Delta\cdot\delta,\tuple{a,k,\delta\backslash\vareps}\cdot\Sigma]}
         {[i,j,\Delta\cdot\gamma,\tuple{a,j,\gamma\backslash\vareps}\cdot\Sigma]}
      {\scriptstyle \gamma \rightarrow \delta \ \gamma\backslash\delta \in P}$,
   \end{center}

   Two rules to invoke left and right function application
   to a fresh component:
   \begin{center}
   $\frac{[i,k,\gamma/\delta,\tuple{i,k,\gamma/\delta}]
          \ [k,j,\Delta\cdot\gamma/\delta\cdot\delta,\Sigma]}
         {[i,j,\Delta\cdot\gamma,\tuple{i,k,\gamma/\delta}\cdot\Sigma]}
     {\scriptstyle \gamma \rightarrow \gamma/\delta \ \delta \in P}$,

   $\frac{[k,j,\gamma\backslash\delta,\tuple{k,j,\gamma\backslash\delta}] 
          \ [i,k,\Delta\cdot\gamma\backslash\delta\cdot\delta,\Sigma]}
         {[i,j,\Delta\cdot\gamma,\tuple{k,j,\gamma\backslash\delta}\cdot\Sigma]}
     {\scriptstyle \gamma \rightarrow \delta \ \gamma\backslash\delta \in P}$,
   \end{center}

   Two rules to discharge empty components:
   \begin{center}
   $\frac{[i,j,\Delta\cdot\gamma/\delta \cdot \delta,\Sigma]}
         {[i,j,\Delta\cdot\gamma,\Sigma]}$
\ 
   $\frac{[i,j,\Delta\cdot\gamma\backslash\delta \cdot \delta,\Sigma]}
         {[i,j,\Delta\cdot\gamma,\Sigma]}$
   \end{center}

   Three rules to skip conjunctions, by adding a gap between the 
   components in a constituent (the first rule consumes the conjunction 
   to create a partial result of category $\mathrm{Conj}'_\delta$, and the 
   latter two use this to skip the opposing NP):
   \begin{center}
   $\frac{[k,j,\Delta\cdot\delta,\Sigma]}
         {[i,j,\Delta\cdot\mathrm{Conj}'_\delta,\Sigma]}
      {\scriptstyle [i,k,\mathrm{Conj},\tuple{i,k,\mathrm{Conj}}]}$

   $\frac{[k,j,\Delta\cdot\mathrm{Conj}'_\delta,\Sigma]}
         {[i,j,\Delta\cdot\delta,\Sigma]}
      {\scriptstyle [i,k,\Delta\cdot\delta,\wild]}$
\
   $\frac{[i,k,\Delta\cdot\delta,\Sigma]}
         {[i,j,\Delta\cdot\delta,\Sigma]}
      {\scriptstyle [k,j,\Delta\cdot\mathrm{Conj}'_\delta,\wild]}$
   \end{center}

   Two rules to reassemble discontinuous constituents (again, using a 
   partial result $\mathrm{Conj}'_\delta$ to reduce the number of ranging variables):
   \begin{center}
   $\frac{[a,c,\mathrm{Conj},\tuple{a,c,\mathrm{Conj}}]
           \ [i,j,\gamma,\Sigma\cdot\tuple{c,b,\delta}]}
         {[i,j,\gamma,\Sigma\cdot\tuple{a,b,\mathrm{Conj}'_\delta}]}$

   $\frac{[i,j,\gamma,\Sigma\cdot\tuple{c,b,\mathrm{Conj}'_\delta}]
           \ [i,j,\gamma,\Sigma\cdot\tuple{a,c,\delta}]}
         {[i,j,\gamma,\Sigma\cdot\tuple{a,b,\delta}]}$
   \end{center}

   Two rules to combine adjacent components:
   \begin{center}
   $\frac{[i,j,\gamma,\Sigma\cdot\tuple{a,c,\delta/\vareps}\cdot\tuple{c,b,\vareps}]}
         {[i,j,\gamma,\Sigma\cdot\tuple{a,b,\delta}]}$
\ 
   $\frac{[i,j,\gamma,\Sigma\cdot\tuple{c,b,\delta\backslash\vareps}\cdot\tuple{a,c,\vareps}]}
         {[i,j,\gamma,\Sigma\cdot\tuple{a,b,\delta}]}$
   \end{center}

   And one rule to apply quantifier functions:
   \begin{center}
   $\frac{[i,j,\gamma,\Sigma\cdot\tuple{a,b,\delta_q}]}
         {[i,j,\gamma,\Sigma\cdot\tuple{a,b,\delta_\epsilon}]}$
   \end{center}

\end{itemize}

\begin{figure*}
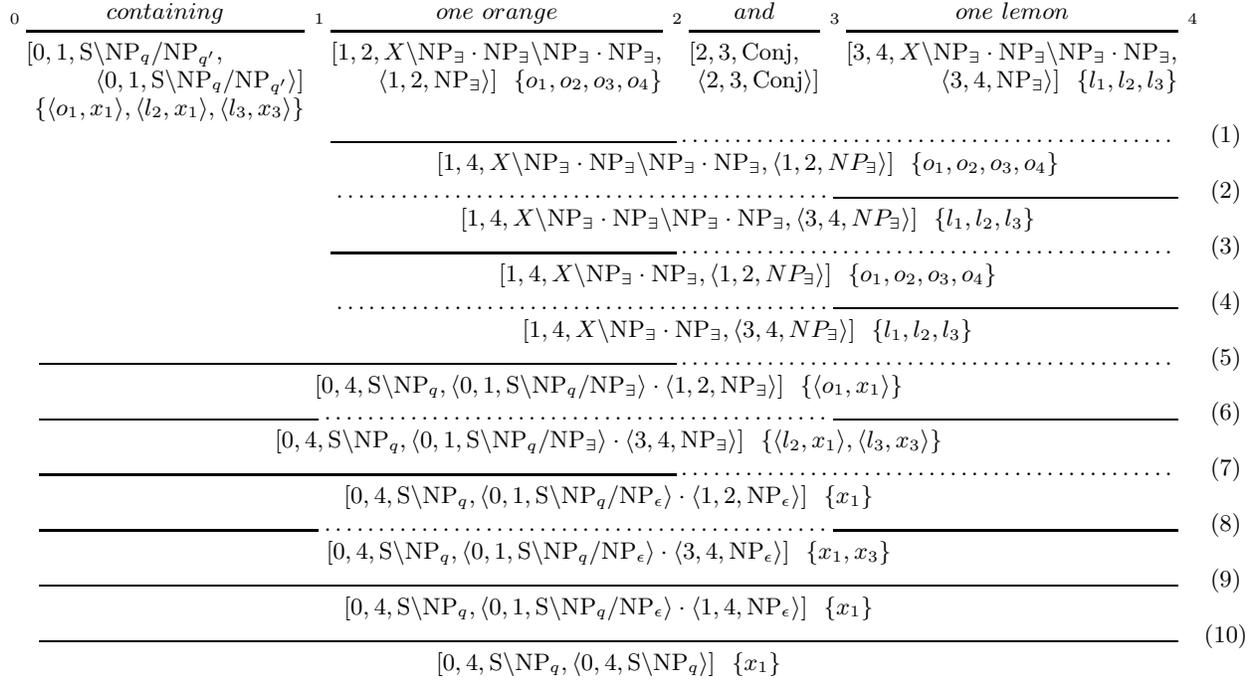

\begin{center}
\small
\(
\begin{array}{@{}ccccc@{}}
\!\!\!\!_0 \hfill & \!\!\!\!_1 \hfill & \!\!\!\!_2 \hfill & 
  \!\!\!\!_3 \hfill & \!\!\!\!_4 \hfill \vspace{-4mm} \\ 
containing & one\ orange & and & one\ lemon                      & \vspace{-2mm} \\ 
\hrulefill & \hrulefill & \hrulefill & \hrulefill \\ 
$[$ 0,1,\mathrm{S}\backslash\mathrm{NP}_q/\mathrm{NP}_{q'}, \hfill   &
  $[$ 1,2, X\backslash\mathrm{NP}_\exists \cdot
      \mathrm{NP}_\exists\backslash\mathrm{NP}_{\exists} \cdot
        \mathrm{NP}_\exists,                                           &
    $[$ 2,3,\mathrm{Conj}, \hfill                                        &
      $[$ 3,4, X\backslash\mathrm{NP}_\exists \cdot
          \mathrm{NP}_\exists\backslash\mathrm{NP}_{\exists} \cdot
            \mathrm{NP}_\exists,                                           & \\
\ \hfill \tuple{0,1,\mathrm{S}\backslash\mathrm{NP}_q/\mathrm{NP}_{q'}} $]$ &
  \ \hfill \tuple{1,2,\mathrm{NP}_\exists} $]$ \ \ \{o_1,o_2,o_3,o_4\} &
    \ \hfill \tuple{2,3,\mathrm{Conj}} $]$                               &
      \ \hfill \tuple{3,4,\mathrm{NP}_\exists} $]$ \ \ \{l_1,l_2,l_3\}     & \\
\ \hfill \{ \tuple{o_1,x_1}, \tuple{l_2,x_1}, \tuple{l_3,x_3} \}     & & & & \\
& \multicolumn{1}{c@{}}{\hrulefill} & \multicolumn{2}{@{}c}{\dotfill}  & (1) \\
& \multicolumn{3}{@{}c}{ $[$ 1,4,
                         X\backslash\mathrm{NP}_\exists \cdot
                           \mathrm{NP}_\exists\backslash\mathrm{NP}_{\exists} \cdot
                             \mathrm{NP}_\exists, \tuple{1,2,NP_\exists} $]$
                         \ \ \{ o_1,o_2,o_3,o_4 \} }                   & \\ 
& \multicolumn{2}{c@{}}{\dotfill} & \multicolumn{1}{@{}c}{\hrulefill}  & (2) \\
& \multicolumn{3}{@{}c}{ $[$ 1,4,
                         X\backslash\mathrm{NP}_\exists \cdot
                           \mathrm{NP}_\exists\backslash\mathrm{NP}_{\exists} \cdot
                             \mathrm{NP}_\exists, \tuple{3,4,NP_\exists} $]$
                         \ \ \{ l_1,l_2,l_3 \} }                       & \\ 
& \multicolumn{1}{c@{}}{\hrulefill} & \multicolumn{2}{@{}c}{\dotfill}  & (3) \\
& \multicolumn{3}{@{}c}{ $[$ 1,4,
                         X\backslash\mathrm{NP}_\exists \cdot \mathrm{NP}_\exists, 
                           \tuple{1,2,NP_\exists} $]$
                         \ \ \{ o_1,o_2,o_3,o_4 \} }                   & \\ 
& \multicolumn{2}{c@{}}{\dotfill} & \multicolumn{1}{@{}c}{\hrulefill}  & (4) \\
& \multicolumn{3}{@{}c}{ $[$1,4,
                         X\backslash\mathrm{NP}_\exists \cdot \mathrm{NP}_\exists, 
                           \tuple{3,4,NP_\exists} $]$
                         \ \ \{ l_1,l_2,l_3 \} }                       & \\ 
\multicolumn{2}{c@{}}{\hrulefill} & \multicolumn{2}{@{}c}{\dotfill}    & (5) \\
\multicolumn{4}{c}{ $[$ 0,4, \mathrm{S}\backslash\mathrm{NP}_q,
                    \tuple{0,1,\mathrm{S}\backslash\mathrm{NP}_q/\mathrm{NP}_\exists}
                      \cdot \tuple{1,2,\mathrm{NP}_\exists} $]$
                    \ \ \{ \tuple{o_1,x_1} \} }                        & \\
\multicolumn{1}{c@{}}{\hrulefill} & \multicolumn{2}{@{}c@{}}{\dotfill} &
                                    \multicolumn{1}{@{}c}{\hrulefill}  & (6) \\
\multicolumn{4}{c}{ $[$ 0,4, \mathrm{S}\backslash\mathrm{NP}_q,
                    \tuple{0,1,\mathrm{S}\backslash\mathrm{NP}_q/\mathrm{NP}_\exists}
                      \cdot \tuple{3,4,\mathrm{NP}_\exists} $]$
                    \ \ \{ \tuple{l_2,x_1}, \tuple{l_3,x_3} \} }       & \\
\multicolumn{2}{c@{}}{\hrulefill} & \multicolumn{2}{@{}c}{\dotfill}    & (7) \\
\multicolumn{4}{c}{ $[$ 0,4, \mathrm{S}\backslash\mathrm{NP}_q,
                    \tuple{0,1,\mathrm{S}\backslash\mathrm{NP}_q/\mathrm{NP}_\epsilon}
                      \cdot \tuple{1,2,\mathrm{NP}_\epsilon} $]$
                    \ \ \{ x_1 \} }                                    & \\
\multicolumn{1}{c@{}}{\hrulefill} & \multicolumn{2}{@{}c@{}}{\dotfill} &
                                    \multicolumn{1}{@{}c}{\hrulefill}  & (8) \\
\multicolumn{4}{c}{ $[$ 0,4, \mathrm{S}\backslash\mathrm{NP}_q,
                    \tuple{0,1,\mathrm{S}\backslash\mathrm{NP}_q/\mathrm{NP}_\epsilon}
                      \cdot \tuple{3,4,\mathrm{NP}_\epsilon} $]$
                    \ \ \{ x_1, x_3 \} }                               & \\
\multicolumn{4}{c}{\hrulefill}                                         & (9) \\
\multicolumn{4}{c}{ $[$ 0,4, \mathrm{S}\backslash\mathrm{NP}_q,
                    \tuple{0,1,\mathrm{S}\backslash\mathrm{NP}_q/\mathrm{NP}_\epsilon}
                      \cdot \tuple{1,4,\mathrm{NP}_\epsilon} $]$
                    \ \ \{ x_1 \} }                                    & \\
\multicolumn{4}{c}{\hrulefill}                                         & (10)\\
\multicolumn{4}{c}{ $[$ 0,4, \mathrm{S}\backslash\mathrm{NP}_q,
                    \tuple{0,4,\mathrm{S}\backslash\mathrm{NP}_q} $]$
                    \ \ \{ x_1 \} }                                    & \\
\end{array}
\)
\end{center}

\caption{Sample derivation of conjoined NP.}
\label{fig:conjderiv}
\end{figure*}

The parsing and scoring functions remain identical to those in Section~\ref{sect:basic},
but an additional $k=1$ case containing a modified projection function $\pi$
is now added to the interpretation function,
in order to make the denotations of quantified constituents 
depend on their associated quantifiers:
\[
D(x) = \!\!\!\!\!\!\!\!\!\!\!
  \underset{a_1\dots\, a_k\, s.t.\, \frac{a_1\dots\, a_k}{x}}{\bigcup} \!
    \begin{cases} R(x)           & \text{if } k=0 \vspace{2mm} \\
                  \pi_{q} D(a_1) & \text{if } k=1 \text{ and } \\
                                 & \ \, \frac{a_1}{x} = 
                                     \frac{[...\tuple{...\delta_q}]}
                                          {[...\tuple{...\delta_\epsilon}]} \\
                  \! \operatornamewithlimits{\bigjoin}\limits_{i=1}^{k} \!
                       D(a_i)    & \text{otherwise} \end{cases}
\]
The modified projection function evaluates a quantifier function $q$ 
on some argument denotation $A$, comparing the cardinality of the 
image of the restrictor set in $A$ with the the cardinality of image of 
the intersected restrictor and body sets in $A$:%
\footnote{Following Keenan and Stavi \shortcite{keenanstavi86}.}
\[
\begin{array}{@{}l@{\,}l@{}}
\pi_q\,A = \{ \tuple{e_2...e_a,t} \mid & 
                  \tuple{\wild,e_2...e_a,\wild}\!\in\!A, \ t = q(|R|,|S|) \\
                & R = A\!\Join\!\{\tuple{\wild,e_2...e_a,\wild}\}, \\
                & S = A\!\Join\!\{\tuple{\wild,e_2...e_a,\!\mathrm{TRUE}}\} \,\}
\end{array}
\]

This algorithm parses a categorial grammar in the
usual way -- constituents are initially added to the chart as single
components covering a certain yield in the input string (the indices
of the component are the same as the indices of the constituent itself), and
they are combined by concatenating the yields of smaller constituents
to make larger ones -- until a conjunction is encountered.
When a conjunction is encountered immediately to the left or right of
a recognized constituent constituent~$x$, and another constituent of the
same category is found immediately beyond that conjunction, the parser
creates a new constituent that has the combined yield of both constituents, 
but copies $x$'s component yield (the string indices of $x$'s original
components) with no change.
This has the effect of creating two new constituents every time two 
existing constituents are conjoined: each with a different component yield, 
but both with the same (combined) constituent yield.
These new discontinuous constituents (with component yields that do not exhaust
their constituent yields) are still treated as ordinary constituents by the
parser, which combines them with arguments and modifiers until all of
their argument positions have been successfully discharged, at which
point pairs of discontinuous constituents with the same constituent yield can be 
reassembled into whole -- or at least less discontinuous -- constituents again.

A sample derivation for the verb phrase \sent{containing one orange and
one lemon,} involving conjunction of existentially quantified noun
phrases, is shown in Figure~\ref{fig:conjderiv}, using the above parse
rules and the lexicalized grammar: 
\[
\begin{array}{r@{\ }c@{\ }l}
\text{containing}    &:& \mathrm{S}\backslash\mathrm{NP}_q/\mathrm{NP}_{q'} \\
\text{one}           &:& X\backslash\mathrm{NP}_q \cdot
                           \mathrm{NP}_q\backslash\mathrm{NP}_q \cdot
                             \mathrm{NP}_q/\mathrm{NP}_\epsilon  \\
                     & & X/\mathrm{NP}_q \cdot
                           \mathrm{NP}_q\backslash\mathrm{NP}_q \cdot
                             \mathrm{NP}_q/\mathrm{NP}_\epsilon  \\
\text{orange, lemon} &:& \mathrm{NP}_\epsilon  \\
\text{and}           &:& \mathrm{Conj}  \\
\end{array}
\]
First the parser applies the {\em skip conjunction} rules to obtain the 
discontinuous constituents shown after steps (1) and (2),
and a component is discharged from each of the resulting constituents 
using the {\em empty component} rule in steps (3) and (4). 
The constituents resulting from (3) and (4) are then composed with the verb constituent 
for \sent{containing} in steps (5) and (6), using the 
{\em left attachment rule for fresh components}.
The quantifiers are then applied 
in steps (7) and (8),
and the resulting constituents are reassembled using the {\em conjunction rules} 
in step (9). 
The adjacent components in the constituent resulting from step (9) are then merged 
using the {\em combination rule} in step (10), producing a complete 
gapless constituent for the entire input.

Since the parser rules are fixed, and the number of components in any
chart constituent is bounded by the maximum number of components in a
category (inasmuch as the rules can only add a component to the
recognized list by subtracting one from the unrecognized list), the
algorithm must run in polynomial space and time on the length of the input
sentence.
Since the cardinality of each constituent's denotation is bounded
by $|{\mathcal E}|^v$ (where ${\mathcal E}$ is the set of entities 
in the environment and $v$ is the maximum valency of any category),
the algorithm runs in worst-case polynomial space on $|{\mathcal E}|$;
and since there is no more than one set composition operation performed 
when a rule is applied, and each composition operation runs in worst-case
quadratic time on the size of its composed sets (due to the quantifier operation), 
the algorithm runs in worst-case polynomial time on $|{\mathcal E}|$ as well.

\section{Evaluation}

The extended parser described above has been implemented and evaluated 
on a corpus of 340 spoken instructions to simulated 
human-like agents in a controlled 3-D environment 
(that of children running a lemonade stand, which was 
deemed suitably familiar to undergraduate student subjects).
The parser was run on the word lattice output of an off-the-shelf speech recognizer 
(CMU Sphinx II) and the parser chart was seeded with every hypothesized word.
The parser was also compared with the recognizer by itself, 
in order to determine the degree to which an environment-based 
approach could complement corpus-based disambiguation.
The systems were evaluated as word recognizers (i.e.~ignoring the brackets in 
the parser output) on the first 100 sentences of the corpus 
(corresponding to the first seven of 33 subjects); 
the latter 240 sentences were reserved for training the recognizer and 
for developing the grammar and semantic lexicon.

The average utterance length was approximately three seconds (subsuming about 
300 frames or positions in the parser chart), containing an average of nine words.
Parsing time averaged under 40 seconds per sentence on a P4-1500MHz, 
most of which was spent in forest construction rather 
than denotation calculation.

Accuracy results show that the parser was able to correctly identify a significant 
number of words that the recognizer missed (and vice versa), such that a 
perfect synthesis of the two (choosing the correct word if it is recognized 
by either system) would produce an average of 8 percentage points more recall 
than the recognizer by itself on successful parses,
and as much as 19 percentage points more for some subjects:%
\footnote{Successful parses are those that result in one or more complete analyses 
of the input, even if the correct tree is not among them.}
%
\begin{center}
\small
\begin{tabular}{c|cc|ccc|c}
        &\multicolumn{2}{c|}{recognizer}&\multicolumn{3}{c|}{parser}& joint  \\
subject &      prec     &     recall    &  fail  &  prec  &  recall & recall \\
\hline
   0    &      76       &      79       &   18   &   72   &   74    &   92   \\
   1    &      77       &      75       &   28   &   63   &   55    &   83   \\
   2    &      70       &      71       &   33   &   49   &   54    &   69   \\
   3    &      71       &      67       &   43   &   49   &   45    &   69   \\
   4    &      66       &      54       &   37   &   44   &   39    &   67   \\
   5    &      53       &      52       &   54   &   36   &   31    &   72   \\
   6    &      84       &      84       &   50   &   56   &   63    &   83   \\
\hline
  all   &      68       &      67       &   37   &   53   &   50    &   75   \\
\end{tabular}
\end{center}
which indicates that the environment may offer a useful additional source 
of information for disambiguation.
Though it may not be possible to implement a perfect synthesis of the 
environment-based and corpus-based approaches, if even half of the above 
gains can be realized, it would mark a significant advance.

\section{Conclusion}

This paper has described an extension to an environment-based parsing
algorithm, increasing its semantic coverage to include quantifier
and conjunction operations without destroying its polynomial worst-case complexity.
Experiments using an implementation of this algorithm on a corpus of 
spoken instructions indicate that 1) the observed complexity of the algorithm 
is suitable for practical user interface applications, and 
2) the ability to draw on 
this kind of environment information in an interfaced application 
has the potential to greatly improve recognition accuracy in 
speaker-independent mixed-initiative interfaces.


\small



\end{document}